# Towards Empathetic Human-Robot Interactions

Pascale Fung, Dario Bertero, Yan Wan, Anik Dey, Ricky Ho Yin Chan, Farhad Bin Siddique, Yang Yang, Chien-Sheng Wu, Ruixi Lin

Human Language Technology Center
Department of Electronic and Computer Engineering
The Hong Kong University of Science and Technology
Clear Water Bay, Hong Kong
`pascale@ece.ust.hk`

**Abstract.** Since the late 1990s when speech companies began providing their customer-service software in the market, people have gotten used to speaking to machines. As people interact more often with voice and gesture controlled machines, they expect the machines to recognize different emotions, and understand other high level communication features such as humor, sarcasm and intention. In order to make such communication possible, the machines need an empathy module in them, which is a software system that can extract emotions from human speech and behavior and can decide the correct response of the robot. Although research on empathetic robots is still in the primary stage, current methods involve using signal processing techniques, sentiment analysis and machine learning algorithms to make robots that can 'understand' human emotion. Other aspects of human-robot interaction include facial expression and gesture recognition, as well as robot movement to convey emotion and intent. We propose Zara the Supergirl as a prototype system of empathetic robots. It is a software-based virtual android, with an animated cartoon character to present itself on the screen. She will get 'smarter' and more empathetic, by having machine learning algorithms, and gathering more data and learning from it. In this paper, we present our work so far in the areas of deep learning of emotion and sentiment recognition, as well as humor recognition. We hope to explore the future direction of android development and how it can help improve people's lives.

## 1   Introduction

From science fiction films to novels, humans have always fantasized – or needed – to have an emotional relationship with intelligent machines.

Many people in the society seem to think that the objective of creating intelligent machines is to "imitate humans" or create a new species of "humans". This misunderstanding has led to the irrational fear of "machines taking over humans" by some people. Their reasoning is obvious – if intelligent machines are supposed to imitate humans then as they become more and more human-like they are bound to have humanly desire for power and dominance. It is obvious if one believes in the premises that we are creating machines to "imitate humans". However, this is far from the reality of artificial intelligence research.

Rather than trying to build some Frankenstein surrogate of the human race, the objective of intelligent machine research and development has always been to help humans. As such, even when we build robot "companions" we are working to create health benefits for the elderly or educational benefits for the young.

In the past couple of decades, interactive dialog systems have been designed as software programs either for the desktop, embedded in an enterprise solution, as cloud services, or as mobile applications. They would have a synthesized voice. Since the 1990s, voice interactive designers have tried to make the dialog prompts more natural, and speech synthesis has made great progress to enable computer voice to sound human like. However, such systems remain invisible and virtual. Even after giving these applications names like Siri or Cortana, users remain emotionally indifferent to such systems as if they are merely using an ATM machine for transactions.

One reason behind this might be something that has been studied by human-robot interaction researchers [39]. It is known that physical embodiment of an intelligent system, whether in virtual simulation or in a robotic form, is important for users to feel related and empathize with the system [34].

More importantly though, physical robots, even extremely humanlike androids, seem cold and distant to humans because while they can sometimes be built to look and even sound emotional, they do not recognize or respond to human emotions and intent. Roboticists make great efforts to build robots in anthropomorphic form so that humans can empathize with them [22], and to have embodied cognition [10]. only to find human users disappointed by the lack of reciprocal empathy from these robots.

It follows that we shall embody interactive dialog systems in simulated or robotic forms. It is also important that we give such systems the ability to both recognize human emotions and intent, as well as expressing its own. Before we share our lives with robots, they need to be able to recognize human emotion and intent, through natural language communications, through facial expression and gesture communications.

In this paper, we describe a proposed framework for building a robotic interactive system with an "empathy module". In Section 2, we describe the design of a prototype empathetic virtual robot system. In Section 3 we describe the personality analysis of our system and in Section 4 the need for the system to handle user challenges to our empathetic interactive virtual robot. To enable different features described in Sections 2 to 4, we need speech recognition, emotion and sentiment recognition from audio and text. We describe our current approaches in these areas in Sections 5 to 6. In Section 5 we first present a brief over view of different deep learning architectures. Section 6 describes our current approach of hybrid HMM-DNN speech recognition system for interactive systems. Section 7 describes our approach of emotion recognition from audio with and without feature engineering. We then discuss sentiment recognition from speech and text in Section 8 with the special case of humor recognition in conversations in Section 9. We summarize and discuss future work in Section 10.

## 2  Architecture of An Empathetic Human-Robot Interactive Platform

To achieve human-machine empathy towards each other, we propose a platform that will consist of the following features and functionalities:

1. Embodiment of the system on a virtual robot platform for human empathy;
2. Emotion and intent expression by the robot;
3. Facial recognition of user ethnicity, age, etc.;
4. Speech recognition of what the user is saying including humor;
5. Natural language understanding of user intent and sentiment including humor;
6. Emotion recognition from facial expressions, speech, audio and language;

Research has shown that humans prefer interacting with machines that are anthropomorphous. So the very first step in designing interactive intelligent systems is to give it a humanlike form.

In recent years, we have seen more efforts to give interactive systems a "face". For example, the Microsoft Tay is a chatbot with a human face and personality. "She" is designed as a teenage female character and exhibits the language model of a typical American teenage girl. The Boston company Jibo made a social robot of that name with animated expressions and movement. Jibo has had a huge success online. Pepper is another social robot that has the capability of recognizing human emotion and sentiments, available in Japan. Pepper also has an anthropomorphous body with a head and body movement designed to express emotion and intent.

For a prototype system, we gave our platform a human name and a cartoon figure. We call it Zara the Supergirl, with a multiracial face and a cartoon body. It was shown at the World Economic Forum in Dalian in September 2015. Her cartoon facial expressions and body movements are template-based and programmed to represent her recognition of user emotion or sentiment. Her lips move when she is talking. Her body gestures are program to express a combination of friendliness, power, knowledge and other emotions she might have while interacting with humans.

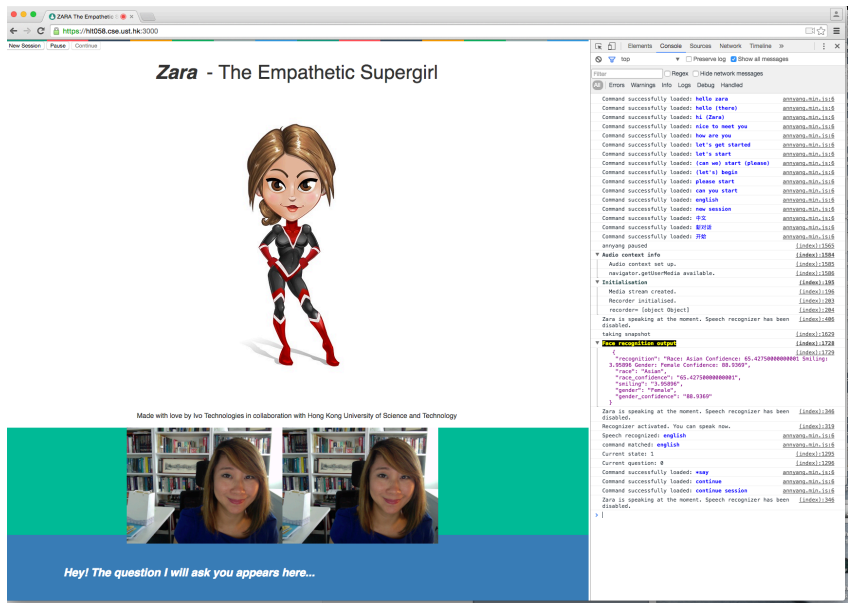

**Fig. 1.** A Prototype Interactive System with Multicue and Emotion Recognition

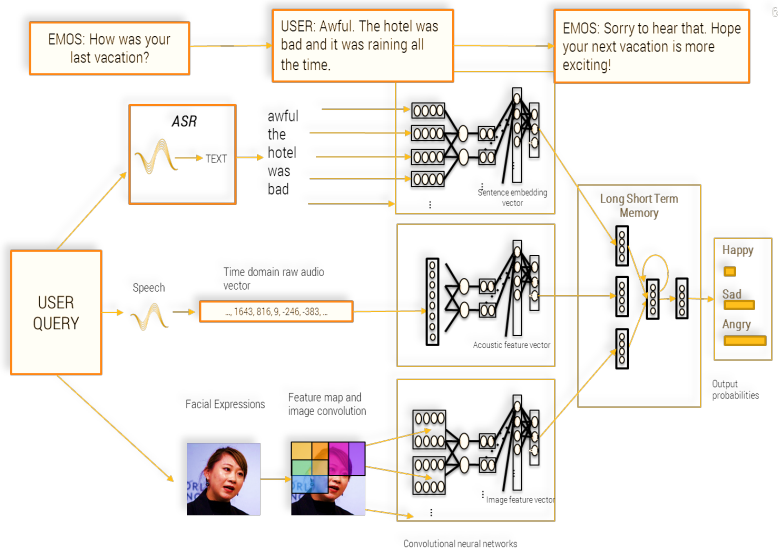

**Fig. 2.** A Deep Learning Multichannel Emotion Recognition Architecture

When you begin a conversation with Zara, she says, "Please wait while I analyze your face". Zara's algorithms study images captured by the computer's webcam to determine your gender and ethnicity. She will then guess which language you speak (Zara understands English, Mandarin and is learning French) and ask you a few questions in your native tongue. What is your earliest memory? Tell me about your mother. How was your last vacation? Through this process, based on your facial expressions, the acoustic features of your voice, and the content of your responses, Zara will respond in ways that mimic empathy. She would frown and express sympathy if you mentioned a sad childhood story, but would give a thumbs up when you talk about a great vacation.

Zara's current task is a conversational MBTI personality assessor and we designed 6 categories of personality-assessing questions, in order to assess the user's personality [24]. After five minutes of conversation, Zara will try to guess your personality—You seem like a easy going and popular person—and ask you about your attitudes toward empathetic machines: How do you feel about machines like me? This is a way for us to gather feedback from people on their interactions with early empathetic robots.

A dialog management system with different states is designed to control the flow of the conversation, which consists of one-part machine-initiative questions from Zara and answers from human users and another part of user-initiative questions and challenges to Zara.

## 3  Personality analysis

Empathy is the recognition and sharing of the emotion of the other. In order to demonstrate machine empathy towards humans within a short duration of interaction, we gave Zara the task of personality analysis. We designed a set of personal questions in six different domains in order to classify user personality from among sixteen different MBTI personality types [24]. The original MBTI test questionnaire contains about 70 questions about user's preferences and sentiments, and would require about half an hour to complete. We asked a group of training users to answer this questionnaire but also answer questions from Zara. The personality type generated by the MBTI questionnaire is used as the gold standard label for training the Zara system. Based on user answers to Zara's questions, scores are calculated in four dimensions (namely Introversion - Extroversion, Intuitive - Sensing, Thinking - Feeling, Judging - Perceiving).

We use the output of the sentiment analysis from language and emotion recognition from speech as linguistic and speech cues to calculate the score for each personality dimension based on previous research [21]. For each response, the individual score for each of the four dimensions is calculated and updated, and the final score in each dimension is the group average of all the responses.

| Level | Introvert | Extravert |
|---|---|---|
| Conversational behaviour | Listen<br>Less back-channel behaviour | Initiate conversation<br>More back-channel behaviour |
| Topic selection | Self-focused<br>Problem talk, dissatisfaction<br>Strict selection<br>Single topic<br>Few semantic errors<br>Few self-references | Not self-focused*<br>Pleasure talk, agreement, compliment<br>Think out loud*<br>Many topics<br>Many semantic errors<br>Many self-references |
| Style | Formal<br>Many hedges (tentative words) | Informal<br>Few hedges (tentative words) |
| Syntax | Many nouns, adjectives, prepositions (explicit)<br>Elaborated constructions<br>Many words per sentence<br>Many articles<br>Many negations | Many verbs, adverbs, pronouns (implicit)<br>Simple constructions*<br>Few words per sentence<br>Few articles<br>Few negations |
| Lexicon | Correct<br>Rich<br>High diversity<br>Many exclusive and inclusive words<br>Few social words<br>Few positive emotion words<br>Many negative emotion words | Loose*<br>Poor<br>Low diversity<br>Few exclusive and inclusive words<br>Many social words<br>Many positive emotion words<br>Few negative emotion words |
| Speech | Received accent<br>Slow speech rate<br>Few disfluencies<br>Many unfilled pauses<br>Long response latency<br>Quiet<br>Low voice quality<br>Non-nasal voice<br>Low frequency variability | Local accent*<br>High speech rate<br>Many disfluencies*<br>Few unfilled pauses<br>Short response latency<br>Loud<br>High voice quality<br>Nasal voice<br>High frequency variability |

**Fig. 3.** Summary of identified language cues for extraversion and various production levels [21]

## 4   Handling user challenge

The personality test consists mostly of machine-initiative questions from Zara and human answers. However, as described in the user analysis section below, there are scenarios where the user does not respond to questions from Zara directly. 24.62% of the users who tried Zara exhibited some form of verbal challenge in their responses during the dialogue conversation, of which 37.5% of users evade the questions with an irrelevant answer. 12.5% of users challenged Zara's ability more directly with questions unrelated to the personality test.

From September to December 2015, a total of 184 responses were recorded in total, 24.61% of the data showed some form of challenge during the extent of the conversation. Challenge here refers to user responses that were difficult to handle and impeded the flow of conversation with Zara. They include the following 6 types:

1. Seeking disclosure reciprocity
2. Asking for clarification
3. Avoidance of topic
4. Deliberate challenge of system ability
5. Abusive language

6. Garbage.

Several of the above categories can be observed in human-human interactions. For instance, seeking disclosure reciprocity is not uncommon in human conversations [41].

Responses that revealed some form of avoidance of topic were the most frequent. Avoidance in psychology is viewed as a coping mechanism in response to stress, fear, discomfort, or anxiety [30]. In the dataset collected, two types of avoidance were observed. Users who actively avoid the topic specifically reveal their unwillingness to continue the conversation ("I don't want to talk about it", "I am in no mood to tell you a story Zara"), while users who adopts a more passive strategy had the intent to discontinue the conversation implied ("Let's continue.", "Make it a quick one", "You know...").

Abusive language includes foul, obscene, culturally and socially inappropriate remarks and the like. Currently collected data revealed surprisingly few inappropriate comments such as "get lost now" and "None of your business". These challenges are comparatively mild. Owing to the context of Zara's role as a personality assessor, the reasons here for abuse could be the need to trust the robotic assessor and response to discomfort.

Asking for clarification examples included "Can you repeat?" and "Can you say it again?" Clarification questions observed in this dataset are primarily non- reprise questions as a request to repeat a previous utterance [26].

Deliberate challenge of system ability was also observed. This took the form of direct requests ("Can I change a topic?", "Why can't you speak English?" in the Chinese mode), or statements un- related to the questions asked ("Which one is 72.1 percent?").

Zara is programmed with a gentle but witty personality to handle different user challenges. For example, when abusive language is repeatedly used against her, she asks for an apology after expressing concern for the user's level of stress. If the user asks a general domain question unrelated to the personality test, Zara will try to entertain the question with an answer from a general knowledge database using a search engine API, much like Siri or Cortana. However, unlike these other systems, Zara will not chat indefinitely with the user but will remind the user of their task at hand, namely the personality test. In addition, Zara is also designed to have a sense of humor.

In the following sections, we will describe our current approaches for different modules of the Zara system, namely speech recognition, emotion and mood recognition from audio, and sentiment analysis from speech and text. We will also describe a first ever approach in humor recognition from conversations.

## 5   DNN, CNN and LSTM

In this section we give a general description of the Deep Neural Network architectures we use in the task described. The real power of any DNN is to reduce a set of low level features into a single low-dimensional feature vector, performing the appropriate feature-selection [40].

The first model we use is the Convolutional Neural Network [9], which is useful to obtain a fixed-length vector representation of an utterance, an audio signal or an image. The basic structure of a CNN takes as input one or more low level input vectors $x_{0..N}$. In case of an utterance each word is encoded as a vector, in case of an audio signal each frame, or in alternative the raw audio sample. The input feature vectors are first fed into a first embedding layer to obtain a low dimensional dense vector, given by:

$$x_i^E = f(W_E x_i + b_E)$$

where f is a non-linear function (sigmoid, hyperbolic tangent or rectified linear), and W the parameters to be trained.

A sliding window then moves over the output vectors of the embedding layer (or the raw audio input itself), and another layer is applied to each group of token or audio frame vectors:

$$x_i^C = f(W_C [x_h^E]_{h \in [x-\frac{c}{2}, x+\frac{c}{2}]} + b_C)$$

where c is the size of the convolution window, and [] denotes the concatenation. This operation allows to capture the local context and extract features from it. A max-pooling operation is then applied to extract the most salient features of all the tokens or frames into a single vector for the whole utterance. The max-pooling takes, for each vector element, the maximum value among all output vectors from the convolution:

$$x_i^{MP} = \max_t(x_{t,i}^C)$$

where t represents the token, and i the vector element.

The Long Short Term Memory (LSTM) [15] is instead useful to model time sequences where it is important to remember the past context. It is an improvement over the Recurrent Neural Network aimed to enhance its memory capabilities. In a standard RNN the hidden memory layer is updated through a function of the input and the hidden layer at the previous time instant:

$$\boldsymbol{h}_t = \tanh(\boldsymbol{W}_x \boldsymbol{x}_t + \boldsymbol{W}_h \boldsymbol{h}_{t-1} + \boldsymbol{b})$$

Where **x** is the network input and **b** the bias term.

This kind of connection is not very effective to maintain the information stored for long time instants, and it does not allow to forget unneeded information between two time steps.

The LSTM enhances the RNN with a series of three multiplicative gates. The structure is the following:

$$\boldsymbol{i}_t = \sigma(\boldsymbol{W}_{i_x} \boldsymbol{x}_t + \boldsymbol{W}_{i_h} \boldsymbol{h}_{t-1} + \boldsymbol{b}_i)$$
$$\boldsymbol{f}_t = \sigma(\boldsymbol{W}_{f_x} \boldsymbol{x}_t + \boldsymbol{W}_{f_h} \boldsymbol{h}_{t-1} + \boldsymbol{b}_f)$$
$$\boldsymbol{o}_t = \sigma(\boldsymbol{W}_{o_x} \boldsymbol{x}_t + \boldsymbol{W}_{o_h} \boldsymbol{h}_{t-1} + \boldsymbol{b}_o)$$
$$\boldsymbol{s}_t = \tanh(\boldsymbol{W}_{s_x} \boldsymbol{x}_t + \boldsymbol{W}_{s_h} \boldsymbol{h}_{t-1} + \boldsymbol{b}_s)$$
$$\boldsymbol{c}_t = \boldsymbol{f}_t \odot \boldsymbol{c}_{t-1} + \boldsymbol{i}_t \odot \boldsymbol{s}_t$$
$$\boldsymbol{h}_t = \tanh(\boldsymbol{c}_t) \odot \boldsymbol{o}_t$$

where ⊙ is the element-wise product. The gate factors **i**, **f**, **o** are able to let through or suppress a specific update contribution, thus allowing a selective information retaining. In this way a cell value can be retained for multiple time steps when **i** = 0, ignored in the output when **o** = 0, and forgotten when **f** = 0.

# 6 Speech Recognition

We experiment with an automatic speech recognition system implemented in house. ASR systems consist of both an acoustic model and a language model in a Bayesian framework.

$$\hat{W} = \arg\max_{W} P(A|W)P(W)$$

For training acoustic models, we train a GMM-HMM for predicting the hidden states and use a Deep Neural Network (DNN) to predict the emission probabilities of the HMM states. We train maximum likelihood (ML) Gaussian mixture Hidden Markov Model (GMM-HMM) with 8000 tied triphone states and 240K Gaussians, using linear discriminant analysis (LDA) and maximum likelihood linear transform (MLLT) feature transformations. We apply boosted maximum mutual information (bMMI) discriminative training on ML trained HMMs.

We train DNN for the emission probabilities of the HMMs with 6 hidden layers, and each hidden layer has 1024 neurons. The DNN is initialized with stacked restricted Boltzmann machines (RBMs) which are pre-trained in a greedy layerwise fashion. Cross-entropy (CE) criterion DNN training is first applied on the state alignments produced by the GMM-HMMs. State alignment is then reproduced with the DNN-HMMs, and DNN training with CE criterion is done again. Sequence-discriminative training of DNN with state level minimum Bayes risk (sMBR) criterion is applied by lattice-based approach on the CE trained DNN-HMMs. We train our acoustic models with the Kaldi speech recognition toolkit [25].

Our text data to train the language model contains 88.6M sentences. It comprises the acoustic model training English transcriptions, web crawled news data, web crawled book data, Cantab filtering sentences on Google 1 billion word language modeling benchmark, weather domain and music domain queries expanded from manually designed templates, and common chat queries. We train Wittenbell smoothing interpolated trigram language model (LM) and CE based recurrent neural network (RNN) LM using the SRI-LM toolkit and CUED-RNNLM toolkit respectively.

The ASR decoder performs search on weighted finite state transducer (WFST) graph for trigram LM and generates lattice, and then performs lattice rescoring with RNN LM. The decoder is designed for input audio data that is streamed from TCP/IP or HTTP network protocol, and also performs decoding in real time. The decoder supports simultaneous users by multiple threads and user queue. The English ASR achieves 7.6% word error rate on the combined test set of wsj1[1] "si_dt_05" and "si_dt_20".

---

[1] https://catalog.ldc.upenn.edu/LDC94S13A

Our system is a hybrid DNN-HMM system. In recent years, there has been work on end-to-end models for speech recognition where feature vectors are the input and word sequences are predicted one by one, without an HMM framework. One such model is called End-to-End Attention-based Large Vocabulary Speech Recognition [3]. Recurrent Neural Nets (RNNs) are used to replace the HMMs. Language models are incorporated directly into the RNN decoder. Other approach includes using Deep Convolutional Neural Nets instead of DNNs or GMMs [32]. CNNs have the advantage of being able to model the spectral correlations that exist in speech signals.

In summary, speech recognition has improved in leaps and bounds thanks to deep learning and is able to give us most of what we want to hear from any spoken input. The decoded word sequence is then used for later stage semantic processing, such as emotion, sentiment and intent recognition.

## 7 Real-Time Emotion Recognition from Time-Domain Raw Audio Input

In recent years, we have seen successful systems that gave high classification accuracies on benchmark datasets of emotional speech [21] or music genres and moods [34]. Most of such work consists of two main steps, namely feature extraction and classifier learning. One challenge for most emotion recognition systems from speech is the time needed to extract features from the speech file. Both high and low level features, as described in the following section, are needed so far for emotion and mood recognition from audio. There are close to a 1000 features, a much larger set than the feature set used for speech recognition, that need to be extracted and computed over windows of audio signals. This typically takes a few dozen seconds to do for each utterance, making the response time less than real-time instantaneous, which users have come to expect from interactive systems.

Feature engineering is tedious and time-consuming. It also requires a lot of hand tuning. In order to bypass feature engineering, the current direction is to explore methods that can recognize emotion or mood directly from time-domain audio signals. One approach that has shown great potential is using Convolutional Neural Networks. In the following sections, we compare an approach of using CNN without feature engineering to a method that uses audio features with a SVM classifier.

### 7.1 Dataset

For our experiments on emotion recognition with raw audio, we built a dataset form the TED-LIUM corpus release 2 [31]. It includes 207 hours of speech extracted from 1495 TED talks. We annotated the data with an existing commercial API followed by manual correction. We obtained a total of 90041 segments, divided into the following 11 categories: creative/passionate, criticism/cynicism, defensiveness/anxiety, friendly/warm, hostility/anger, leadership/charisma, loneliness/unfulfillment, love/happiness, sadness/sorrow, self-control/practicality and supremacy/arrogance.

In our experiments we only use the following 6 categories: criticism, anxiety, anger, loneliness, happiness, sadness. We obtained a total of 2389 segments for the criticism category, 3855 for anxiety, 12708 for anger, 3618 for loneliness, 8070 for happy and 1824 for sadness. The segments have an average length slightly above 13 seconds. As on-going work, we continue to re-label audio data with our CNN-based emotion decoder followed by manual correction.

## 7.2  Real-Time Speech Emotion Recognition with CNN

The CNN model using raw audio as input is shown in Figure 4. The raw audio samples are first down-sampled at 8kHz, in order to optimize between the sampling rate and representation memory efficiency in case of longer segments. The CNN is designed with one filter for real-time processing. We set a convolution window of size 200, which corresponds to 25 ms, and an overlapping step size of 50, equal to around 6 ms. The convolution layer perform the feature extraction in each layer, and models the variations among neighboring frames due to the overlapping.

The network is trained using the standard back-propagation algorithm, performing gradient descent over each parameter. At evaluation time the time complexity is linear over the length of the audio input segment due to the convolution. Thus the largest time contribution is due to the computations inside the network [14], which with one convolution only can be performed in negligible time for single utterances.

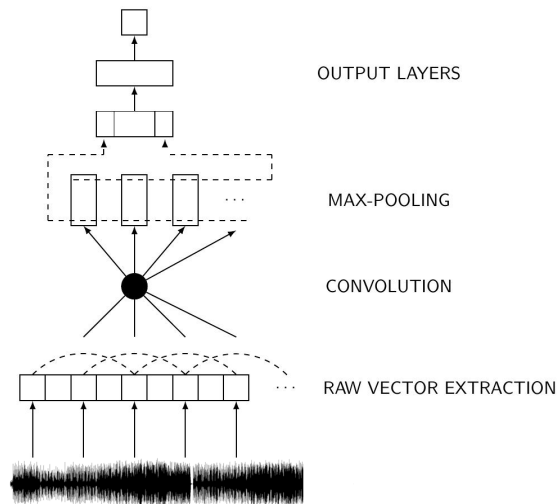

**Fig. 4.** Convolutional Neural Network For Emotion Classification of Raw Audio.

## 7.3 Experimental Setup

We setup our experiments as binary classification tasks, in which each segment is classified as either part of a particular emotion category or not part of that particular category. For each category the negative samples are chosen randomly from the music clips that do not belong to the positive genre.

We implement our CNN with the Theano framework [4]. Theano's automatic differentiation capabilities are used to implement the backpropagation. Our models are trained with GPU Tesla K20 on the CUDA platform.

We choose rectified linear as the non-linear function for the hidden layers, as it generally provides better performance over other functions. We use standard backpropagation training, with momentum set to 0.9 and initial learning rate to $\mathbf{10^{-5}}$. We used the validation set to determine the early stopping condition when the error on it began to increase. We normalize the input data of each experiment with zero mean and unit standard deviation.

As a baseline we use a linear-kernel SVM model from the LibSVM [8] library with the INTERSPEECH 2009 emotion feature set [35], extracted with openSMILE [12]. These features are computed from a series of input frames and output a single static summary vector, e.g, the smooth methods, maximum and minimum value, mean value of the features from the frames [20]. The model of each binary classification is trained through 500 iterations.

All results are shown in table 1, while Figure 5 shows the learning curves of our binary classification using raw audio data as input. The lower results for some categories, even on the SVM baseline, may be a sign of inaccuracy in manual labeling. We plan to work to improve both the dataset, with hand-labeled samples, and retrain the model as on going work.

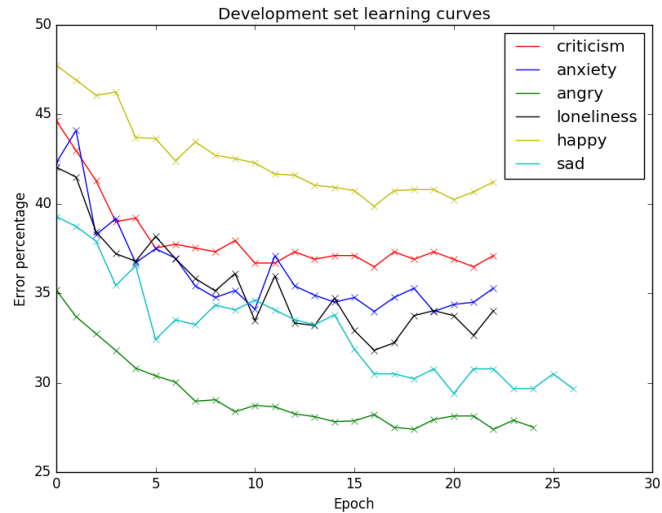

**Fig. 5.** Learning curve for binary classification: Raw audio data (CNN) over six emotion categories.

**Table 1.** Real-Time CNN Outperforms SVM with Features

|  | % Accuracy (CNN) | % Accuracy (SVM) |
|---|---|---|
| Criticism, Cynicism | 61.2 | 55.0 |
| Defensiveness, Anxiety | 62.0 | 56.3 |
| Hostility, Anger | 72.9 | 72.8 |
| Loneliness, Unfulfillment | 66.6 | 61.1 |
| Love, Happiness | 60.1 | 50.9 |
| Sadness, Sorrow | 71.4 | 71.1 |
| **Average** | **65.7** | **61.2** |

## 8    Sentiment Inference From Speech and Text

In the first version of Zara, sentiment analysis was based on lexical features. We look for keyword matches from a pool of positive and negative emotion lexicons from LIWC[2] dictionary and use an N-gram model to classify the sentiment. In the current approach, we use a CNN-based classifier on Word2Vec. Convolutional Neural Networks (CNNs) have recently achieved remarkably strong performance on the practically important task of sentence classification [16,17,18].

---

[2] http://liwc.wpengine.com

In this work, we train a CNN with one layer of convolution and max pooling on top of word vectors obtained from an unsupervised neural language model. We begin with a sentence that we then convert to a matrix. The rows are word vector representations of each word in that sentence. There are several publicly available word vectors sets, such as Google word2vec, WordNet or GloVe. Due to the instinctive sequential structure of a sentence, we use filters that slide over full rows of the matrix, which scrolls word by word specifically. The width of the filter is fixed to the dimensionality of the word vector, and the height varies with different filters. Our model uses multiple filters (with different window heights 3, 4 and 5) to represent multiple features. The convolution matrix $W_i^{cnv}$ is the parameters to be learned for the $i$-th filter and its size is a hyper-parameter to be chosen for development. In our model, we choose 100. We then apply a max-over-time pooling operation over each feature, which means the model will always pick up the most valuable information wherever it happens in the input sentence. Also, the pooling technique can fix the feature length for further classification. Our work is implemented as a multi-channel model, which has two channels of word vectors—one that is kept static throughout training as word2vec and another is fine-tuned via back propagation. At last, we concatenate the feature outputs of each channel and pass it to a fully connected softmax layer whose output is the probability distribution over a binary classification for sentiment analysis of text transcribed from speech by our speech recognizer.

Table 2. Corpus statistics for text emotion experiments with CNN

| Corpus | Average length | Size | Vocabulary size | Words in Word2vec |
|---|---|---|---|---|
| Movie Review Training Data | 20 | 10662 | 18765 | 16448 |

Table 3. CNN sentiment analysis on 388 sentences from human interacting with Zara

|  | Accuracy | Precision | Recall | F-score |
|---|---|---|---|---|
| CNN | 67.8 % | 91.2% | 63.5% | 74.8% |

It is also possible to combine audio input and transcribed text into the CNNs with two channels, as illustrated in the Figure 6 below. Each channel process either the speech or the text features, through the max-pooling layer, the largest number of each feature is recorded into the next layer. The output of each channel are concatenated to forma a feature vector for the penultimate layer, which will be further encoded for the final layer. The final softmax will be performed for the final layer.

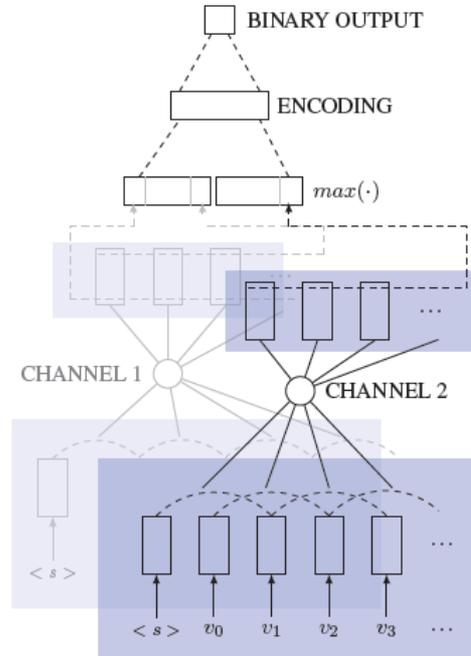

**Figure 6**. Bi-channel CNN for Sentiment Analysis from Speech and Text.

## 9     Humor Recognition in Conversations

In this section, we will describe a fully machine-learning approach for learning a particular sentiment – humor from spoken conversations. The methodology is applicable to recognizing other sentiments as well.

Humor can be classified depending on the way they are intended to trigger laughter: common forms include irony, sarcasm, puns and nonsense [1]. Humor in a conversation serves to lighten the mood and build rapport between the speaker and the audience. It is also very helpful for human-robot interactions. Humans are more forgiving of machine mistakes when the latter has a sense of humor. Yell at Siri, you will get a "Hey! I am doing my best!". Ask a challenging question like "Siri, do you love me?" the answer is likely to be "I am not at capable of love".

    Humor in a spoken dialog is characterized both by what is said and how it is said. Therefore prosody is an important component that must be taken into account together with the semantic content of an utterance. A long-term goal of humor research, or language understanding in general, is how to effectively integrate language features with audio features.

    We analyze funny dialogues extracted from humorous TV-sitcoms. A sitcom is a scripted oral dialog where at regular intervals canned laughter are embedded. Such

laughter indicates where the audience is supposed to laugh, and solicit its active participation. We are interested to predict when this occurs.

Spontaneous conversational humor typically follows a defined recurrent structure [2,38]. The first moment is the "setup", which outlines the context for the subsequent jokes and prepare the audience to receive their stimuli. A specific utterance called "punchline" has then the effect to release the tension with a peculiar reaction, typically laughter. Assuming the punchlines are the utterances followed by a canned laughter, our task is to detect them. To keep the show interesting and the audience engaged, sitcom punchlines are equally distributed throughout the show duration. This makes them a rich source of funny humorous dialogues.

An example of a sitcom dialog is shown below:
PENNY: Okay, Sheldon, what can I get you?
SHELDON: Alcohol.
PENNY: Could you be a little more specific?
SHELDON: **Ethyl alcohol. LAUGH Forty milliliters. LAUGH**
PENNY: I'm sorry, honey, I don't know milliliters.
SHELDON: **Ah. Blame President James Jimmy Carter. LAUGH**
**He started America on a path to the metric system but then just gave up. LAUGH**

In the example the punchlines are highlighted in bold.

Before each punchline the other utterances build the setup. In a dialog setting, without a proper context the punchlines may lose their effect of triggering laughters. If we take the last utterance of the example above out of context it may be perceived as a political complaint. Some people may still laugh if exposed to this utterance alone, but setting the dialog in a bar makes the humorous intent stronger.

We employ a supervised classification method to detect when punchlines occur. We use a Deep Neural Network framework divided into two levels. The first level is made of two Convolutional Neural Networks [9] to encode each individual utterance from word embedding vectors, and the audio track associated from a set of frame-level features. The language CNN is followed by a Long Short-Term Memory [15] to model the sequential context of the dialog. Before the output softmax layer we concatenate the output of the LSTM with the acoustic feature vector of the audio CNN, and a few sentence-based extra features. A framework diagram is shown in Figure 7.

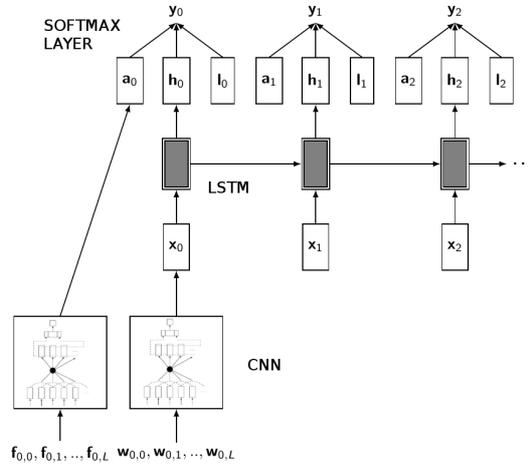

**Fig. 7.** Humor Inference Network : **w** are the word embedding vectors, **f** the audio frames feature vectors, and **l** the other language feature vectors.

**Convolutional Neural Network**

A straightforward way to model a dialog is to retrieve language and acoustic features from each utterance, and to apply a memory-based classifier (such as a Recurrent Neural Network or a Long Short-Term Memory) to model the discourse context. Previous work showed that acoustic utterance-level features (such as the INTERSPEECH 2010 paralinguistic challenge feature set [36]) are quite effective with simple classifiers such as logistic regression and conditional random fields, but yield suboptimal performance with DNN systems [5,13]. We apply two parallel Convolutional Neural Networks to model each utterance from lower-level features.

The first CNN we use is dedicated to language features. Word2Vec word vectors are used as token-level features. Our Word2Vec model is trained on the text9 Wikipedia corpus[3] using gensim implementation [28]. Word2Vec is an improvement over bag-of-ngram or similar representations that ignore out-of-vocabulary words, that often occur in the cross-domain tasks, for example in the case of different sitcoms with different domain vocabulary.

The second CNN is used to encode the audio track of each utterance. We split each utterance into overlapping 25ms frames shifted 10ms from each other. We then extract for each frame a vector of low level acoustic and prosodic features with openSMILE [12]. The features we extract are MFCC, Δ MFCC, ΔΔ MFCC, pitch, energy and zero crossing mean. A CNN then combines together all the frames into a single vector for each utterance, selecting the features from the most salient frames and discarding those that carry no information. This CNN is similar to that used for language, but it uses two embedding layers at the input, and rectified linear units instead of tanh. Both past attempts in the literature [13] and our experiments on our development sets showed the use of frame-level features over a shifting-window, and of rectified linear units to be more effective.

---

[3] Extension of the text8 corpus, obtained from http://mattmahoney.net/dc/textdata

**Long Short-Term Memory for the utterance sequence**

As dialog utterances are sequential, we feed the utterance vectors in sequence into a LSTM block to incorporate contextual information. The memory unit keeps track of the context of previous utterances, and mimics human memory to accumulate the setup that may trigger a punchline.

Before the final softmax layer we incorporate some extra features not modeled by our neural network but which still add important contribution [6,7,27,29].

These features are average word length, sentence length and difference in sentence length with the previous utterances of the context window, speaker identity as well as the speaking rate (time duration of the utterance divided by the sentence length).

All these features are concatenated to the LSTM output together with the audio CNN output, and a softmax layer is applied to get the final output probabilities.

## 9.1 Experiments

We built a corpus from three popular TV-sitcoms: "The Big Bang Theory" seasons 1 to 6, "Friends" seasons 6 to 9, and "Seinfeld" seasons 5 to 9. We downloaded the subtitle files associated to each episode and the scripts[4]. We extracted and segmented the audio track of each episode for the acoustic features. The audio tracks were also used to retrieve the canned laughter timestamps, applying a vocal removal tool followed by a silence/sound detector. We then annotated each utterance as a punch line in case it was followed by laughter within 1s, assuming that utterances not followed by a laughter would be the setup for the punch line. Other than for annotation, we used the timestamps to cut from each utterance audio track the overlapping parts with a canned laughter, in order to avoid crosstalk bias.

For each show we built a training set of 80% of the overall episodes, and a development and test set of 10% each. The episodes were drawn from all the seasons with the same proportion. Our three corpora are structured as follows:

We set the size to 100 for all the hidden layers of the language CNN and the LSTM and the convolution window to 5. In the audio CNN we set the size to 50 and we used a convolution window of size 3. We concatenated the audio CNN output with the other features at the end instead of feeding it through the LSTM, as it gave us better performance. We applied a dropout regularization layer [37] after the output of the LSTM and the audio CNN, the dropout coefficient was set to 0.7. The network was trained with standard back propagation, with a momentum coefficient set to 0.9. The development set was used to tune the hyperparameters, and to determine the early stopping condition. The neural network was implemented with the Theano toolkit [4].

---

[4] From bigbangtrans.wordpress.com and http://www.livesinabox.com/friends/scripts.shtml

In the first set of experiments we trained and tested the classifiers on the same sitcom. As a first baseline for comparison we chose a Conditional Random Field [19] trained over a set of features [6] including the same features added at the end of our neural network, bag-of-ngrams, part of speech proportion, sentiment from SentiWordNet [11], antonyms, and a prosodic feature vector from the INTERSPEECH 2010 paralinguistic challenge [36]. We used the implementation from CRFSuite [23], with L2 regularization. The second baseline system is made of the CNNs encoding blocks only, where the role of the LSTM is replaced with the language CNN output vectors at previous times, concatenated together before the softmax layer. We evaluated context window length of 1 (no LSTM), 2, 3 and 5; as in all the sitcoms over 80% of punchlines occur within five utterances from the previous. All the results are shown in table 4.

Table 4. CNN+LSTM Gives Superior Performance on Humor Recognition

| Context window size | CNN + shifted context | | | | CNN + LSTM | | | |
| --- | --- | --- | --- | --- | --- | --- | --- | --- |
| | A | P | R | F1 | A | P | R | F1 |
| **The Big Bang Theory** | | | | | | | | |
| CRF baseline | 73.4 | 72.1 | 61.8 | 66.5 | 73.4 | 72.1 | 61.8 | 66.5 |
| 1 utterance | **73.7** | 70.3 | 66.7 | 68.5 | **73.7** | 70.3 | 66.7 | 68.5 |
| 2 utterances | 73.0 | 70.3 | 63.8 | 66.9 | 72.8 | 68.0 | 68.7 | 68.4 |
| 3 utterances | 72.2 | 64.9 | 76.2 | 70.1 | 71.6 | 63.6 | **78.7** | **70.3** |
| 5 utterances | 71.2 | 66.7 | 65.4 | 66.0 | 72.9 | **72.6** | 59.1 | 65.2 |
| **Friends** | | | | | | | | |
| CRF baseline | **71.6** | **62.7** | 45.7 | 52.9 | **71.6** | **62.7** | 45.7 | 52.9 |
| 1 utterance | 70.6 | 61.1 | 42.8 | 50.3 | 70.6 | 61.1 | 42.8 | 50.3 |
| 2 utterances | 70.6 | 60.2 | 45.9 | 52.1 | 68.3 | 54.6 | **54.0** | **54.3** |
| 3 utterances | 69.1 | 58.9 | 37.4 | 45.8 | 69.2 | 56.6 | 50.0 | 53.1 |
| 5 utterances | 70.8 | 62.3 | 40.9 | 49.4 | 70.7 | **62.7** | 41.4 | 49.9 |
| **Seinfeld** | | | | | | | | |
| CRF baseline | 76.5 | 59.4 | 36.5 | 45.2 | 76.5 | 59.4 | 36.5 | 45.2 |
| 1 utterance | 71.1 | 44.4 | 35.4 | 39.4 | 71.1 | 44.4 | 35.4 | 39.4 |
| 2 utterances | 72.2 | 46.4 | 30.0 | 36.5 | **76.6** | 65.8 | 25.1 | 36.4 |
| 3 utterances | 70.5 | 43.6 | 38.0 | 40.6 | 75.3 | 54.4 | **44.7** | **49.1** |
| 5 utterances | 74.3 | 56.3 | 14.9 | 23.6 | **76.6** | **67.8** | 22.6 | 33.9 |

In Table 5 we also compare the results of our best system on the Big Bang corpus with the system proposed in [7], where a LSTM is applied to a larger set of language-only features, which includes one-hot word vectors and character-trigram input vectors in addition to Word2Vec. This comparison evaluates the role of acoustic features in our neural network framework.

The second series of experiments is a cross-domain evaluation. Each classifier is trained over all the data from two of the corpus shuffled together, and tested over the third corpus. In these experiments we ignore the speaker identity feature. Our baseline system is again the CRF described above, trained and tested over the same cross-domain data. We evaluated context window lengths of size 3 and 5. The results of this experiments are shown in table 6.

**Table 5.** Bichannel LSTM Performs the Best

| Method | A | P | R | F1 |
|---|---|---|---|---|
| CRF baseline | 65.9 | 61.2 | 55.3 | 58.1 |
| LSTM language-only | 70.0 | 66.7 | 59.4 | 62.9 |
| LSTM language+audio | 71.6 | 63.6 | 78.7 | **70.3** |

**Table 6.** LSTM is Relatively Robust Across Different Corpora

| Train: Seinfeld + Friends | Test: Big Bang | | | | | | | |
|---|---|---|---|---|---|---|---|---|
| | Self-trained | | | | Cross-corpus | | | |
| Context window size | A | P | R | F1 | A | P | R | F1 |
| CRF baseline | 73.4 | 72.1 | 61.8 | 66.5 | 68.3 | 73.3 | 40.5 | 52.2 |
| LSTM 3 utterances | 71.5 | 63.6 | 78.7 | 70.3 | 69.6 | 72.2 | 47.1 | 57.0 |
| LSTM 5 utterances | 72.9 | 72.6 | 59.1 | 65.2 | 68.7 | **76.1** | 39.0 | 51.6 |

| Train: Big Bang + Seinfeld | Test: Friends | | | | | | | |
|---|---|---|---|---|---|---|---|---|
| | Self-trained | | | | Cross-corpus | | | |
| Context window size | A | P | R | F1 | A | P | R | F1 |
| CRF baseline | 71.6 | 62.7 | 45.7 | 52.9 | 63.0 | 47.4 | 56.9 | 51.7 |
| LSTM 3 utterances | 69.2 | 56.6 | 50.0 | 53.1 | 69.6 | 71.0 | 21.4 | 32.9 |
| LSTM 5 utterances | 70.7 | 62.7 | 41.4 | 49.9 | 69.5 | **72.0** | 20.3 | 31.6 |

| Train: Big Bang + Friends | Test: Seinfeld | | | | | | | |
|---|---|---|---|---|---|---|---|---|
| | Self-trained | | | | Cross-corpus | | | |
| Context window size | A | P | R | F1 | A | P | R | F1 |
| CRF baseline | 76.5 | 59.4 | 36.5 | 45.2 | 69.6 | 44.4 | 58.1 | 50.3 |
| LSTM 3 utterances | 75.3 | 54.4 | 44.7 | 49.1 | 64.5 | 40.2 | 74.8 | 52.3 |
| LSTM 5 utterances | 76.6 | 67.8 | 22.6 | 33.9 | 67.8 | 43.2 | 67.7 | **52.7** |

## 10 Summary and Discussion

In this paper, we have described a prototype system of an empathetic virtual robot that can recognize user emotions and thereby bring about a new level of human-robot interactions. We described the design of the architecture, the task of personality analysis, and user analysis of Zara the Supergirl. From there, we extended our description to include more details of the recognition and inference of emotion and sentiment from speech and language. Zara also has a facial recognition component which we have not described in detail as it acts as a supplement to the speech and language part.

We have shown how Deep Learning can be used for various modules in this architecture, ranging from speech recognition, emotion recognition to humor recognition from dialogs. More importantly, we have shown that by using a CNN with one filter, it is possible to obtain real-time performance on speech emotion recognition directly from time-domain audio input, bypassing feature engineering. We have so far developed only the most primary tools that future emotionally intelligent robots would need. Empathetic robots including Zara that are there currently, and the ones that will be there in the near future, might not be completely perfect. However, the most significant step is to make robots to be more human like in their interactions. This means it will have flaws, just like humans do. If this is done right, then future machines and robots will be empathetic and less likely to commit harm in their interactions with humans. They will be able to get us, understand our emotions, and more than anything, they will be our teachers, our caregivers, and our friends.

## References

1. Attardo, S. .Linguistic theories of humor (Vol. 1). Walter de Gruyter. (1994)
2. Attardo, S. The semantic foundations of cognitive theories of humor.*Humor-International Journal of Humor Research*, *10*(4), 395-420. (1997)
3. Bahdanau, D., Chorowski, J., Serdyuk, D., Brakel, P., & Bengio, Y. (2015). End-to-end attention-based large vocabulary speech recognition. In *Acoustics, Speech and Signal Processing (ICASSP), 2016 IEEE International Conference on,* (2016)
4. Bergstra, J., Breuleux, O., Bastien, F., Lamblin, P., Pascanu, R., Desjardins, G., ... & Bengio, Y. Theano: a CPU and GPU math expression compiler. In *Proceedings of the Python for scientific computing conference (SciPy)* (Vol. 4, p. 3). (2010)
5. Bertero. D., Fung, P. Deep learning of audio and language features for humor prediction. *in International Conference on Language Resources and Evaluation (LREC)* (2016)
6. Bertero. D., Fung, P. Predicting humor response in dialogues from TV sitcoms. In *Acoustics, Speech and Signal Processing (ICASSP), 2016 IEEE International Conference on,* (2016)
7. Bertero. D., Fung, P. A long short-term memory framework for predicting humor in dialogues *in Proceedings of the 2016 Conference of the North American Chapter of the Association for Computational Linguistics: Human Language Technologies* (2016)
8. Chang, C. C., & Lin, C. J. LIBSVM: a library for support vector machines. *ACM Transactions on Intelligent Systems and Technology (TIST)*,2(3), 27. (2011)
9. Collobert, R., Weston, J., Bottou, L., Karlen, M., Kavukcuoglu, K., & Kuksa, P. Natural language processing (almost) from scratch. *The Journal of Machine Learning Research*, *12*, 2493-2537. (2011)
10. Duffy, B. R., & Joue, G. Intelligent robots: The question of embodiment. In *Proc. of the Brain-Machine Workshop*. (2000)
11. Esuli, A., & Sebastiani, F. Sentiwordnet: A publicly available lexical resource for opinion mining. In *Proceedings of LREC* (Vol. 6, pp. 417-422). (2006)
12. Eyben, F., Wöllmer, M., & Schuller, B. Opensmile: the munich versatile and fast open-source audio feature extractor. In*Proceedings of the 18th ACM international conference on Multimedia* (pp. 1459-1462). ACM. (2010)
13. Han, K., Yu, D., & Tashev, I. Speech emotion recognition using deep neural network and extreme learning machine. In *Interspeech* (pp. 223-227) (2014)

14. He, K., & Sun, J. (2015). Convolutional neural networks at constrained time cost. In *Proceedings of the IEEE Conference on Computer Vision and Pattern Recognition* (pp. 5353-5360) (2015).
15. Hochreiter, S., & Schmidhuber, J. Long short-term memory. *Neural computation*, *9*(8), 1735-1780. (1997)
16. Johnson, R., & Zhang, T. Effective use of word order for text categorization with convolutional neural networks. *In Proceedings of the 53rd Annual Meeting of the Association for Computational Linguistics*. (2015)
17. Kalchbrenner, N., Grefenstette, E., & Blunsom, P. (2014). A convolutional neural network for modelling sentences. *In Proceedings of the 52nd Annual Meeting of the Association for Computational Linguistics* (2014)
18. Kim, Yoon, Conditional Neural Networks for sentence classification. In *EMNLP* 2014.
19. Lafferty, J., McCallum, A., & Pereira, F. C. Conditional random fields: Probabilistic models for segmenting and labeling sequence data. (2001)
20. Liscombe, J., Venditti, J., & Hirschberg, J. B. Classifying subject ratings of emotional speech using acoustic features. *Columbia University Academic Commons*. (2003)
21. Mairesse, F., Walker, M. A., Mehl, M. R., & Moore, R. K. Using linguistic cues for the automatic recognition of personality in conversation and text. *Journal of artificial intelligence research*, 457-500. (2007).
22. Mataric, M. J. The role of embodiment in assistive interactive robotics for the elderly. In *AAAI fall symposium on caring machines: AI for the elderly, Arlington, VA*. (2005)
23. Okazaki, N. CRFsuite: a fast implementation of conditional random fields (CRFs). *URL http://www. chokkan. org/software/crfsuite*. (2007)
24. Polzehl, T., Möller, S., & Metze, F. Automatically assessing personality from speech. In *Semantic Computing (ICSC), 2010 IEEE Fourth International Conference on* (pp. 134-140). IEEE. (2010)
25. Povey, D., Ghoshal, A., Boulianne, G., Burget, L., Glembek, O., Goel, N. & Silovsky, J. The Kaldi speech recognition toolkit. In *IEEE 2011 workshop on automatic speech recognition and understanding* (No. EPFL-CONF-192584). IEEE Signal Processing Society. (2011)
26. Purver, M. The theory and use of clarification requests in dialogue. *Unpublished doctoral dissertation, University of London*. (2004)
27. Rakov, R., & Rosenberg, A. " sure, i did the right thing": a system for sarcasm detection in speech. In *INTERSPEECH* (pp. 842-846). (2013)
28. Řehůřek, R., & Sojka, P. Gensim–Python Framework for Vector Space Mo delling. *NLP Centre, Faculty of Informatics, Masaryk University, Brno, Czech Republic*. (2011)
29. Reyes, A., Rosso, P., & Veale, T. A multidimensional approach for detecting irony in twitter. *Language resources and evaluation*, *47*(1), 239-268. (2013)
30. Roth, S., & Cohen, L. J. Approach, avoidance, and coping with stress. *American psychologist*, 41(7), 813. (1986)
31. Rousseau, A., Deléglise, P., & Estève, Y. Enhancing the TED-LIUM Corpus with Selected Data for Language Modeling and More TED Talks. In *LREC* (pp. 3935-3939) (2014)
32. Sainath, T. N., Mohamed, A. R., Kingsbury, B., & Ramabhadran, B. Deep convolutional neural networks for LVCSR. In *Acoustics, Speech and Signal Processing (ICASSP), 2013 IEEE International Conference on* (pp. 8614-8618). IEEE. (2013)
33. Scaringella, N., Zoia, G., Mlynek. D. Automatic genre classification of music content: a survey. *Signal Processing Magazine, IEEE, 23(2):133– 141*, (2006)
34. Schermerhorn, P., & Scheutz, M. Disentangling the effects of robot affect, embodiment, and autonomy on human team members in a mixed-initiative task. In *Proceedings from the*

*International Conference on Advances in Computer-Human Interactions* (pp. 236-241). (2011)

35. Schuller, B., Steidl, S., & Batliner, A. The INTERSPEECH 2009 emotion challenge. In *INTERSPEECH* (Vol. 2009, pp. 312-315) (2009)
36. Schuller, B., Steidl, S., Batliner, A., Burkhardt, F., Devillers, L., Müller, C. A., & Narayanan, S. S. (2010, September). The INTERSPEECH 2010 paralinguistic challenge. In *INTERSPEECH* (Vol. 2010, pp. 2795-2798) (2010)
37. Srivastava, N., Hinton, G., Krizhevsky, A., Sutskever, I., & Salakhutdinov, R. Dropout: A simple way to prevent neural networks from overfitting.*The Journal of Machine Learning Research*, *15*(1), 1929-1958. (2014)
38. Taylor, J., & Mazlack, L. Toward computational recognition of humorous intent. In *Proceedings of Cognitive Science Conference* (pp. 2166-2171). (2005)
39. Wainer, J., Feil-Seifer, D. J., Shell, D. A., & Matarić, M. J. The role of physical embodiment in human-robot interaction. In *Robot and Human Interactive Communication, 2006. ROMAN 2006. The 15th IEEE International Symposium on* (pp. 117-122). IEEE. (2006)
40. Wang, M., & Manning, C. D. Effect of Non-linear Deep Architecture in Sequence Labeling. In *IJCNLP* (pp. 1285-1291). (2013)
41. Wheeless, L. R., & Grotz, J. The measurement of trust and its relationship to self-disclosure. *Human Communication Research*, 3(3), 250-257. (1977)